\def\year{2020}\relax
\documentclass[letterpaper]{article} 
\usepackage{aaai20}  
\usepackage{times}  
\usepackage{helvet} 
\usepackage{courier}  
\usepackage[hyphens]{url}  
\usepackage{graphicx} 
\usepackage{color}
\usepackage{amsmath}
\usepackage{amsfonts}
\usepackage{multirow}
\usepackage{diagbox}
\usepackage{graphicx}
\usepackage{gensymb}
\usepackage{url}
\usepackage{graphicx}  
\usepackage{caption}
\usepackage{subcaption}
\usepackage{graphicx}

\def\dname{\textit{V-Granger}}

\title{Estimating Granger Causality with Unobserved Confounders 
via Deep Latent-Variable Recurrent Neural Network}
\author{Yuan Meng\\
meng-y16@mails.tsinghua.edu.cn}
\begin{document}

\maketitle

\begin{abstract}
Granger causality analysis, as one of the most popular time series causality methods, has been widely used in the economics, neuroscience. However, unobserved confounders is a fundamental problem  in the observational studies, which is still not solved for the non-linear Granger causality. The application works often deal with this problem in virtue of the proxy variables, who can be treated as a measure of the confounder with noise. But the proxy variables has been proved to be unreliable, because of the bias it may induce. In this paper, we try to "recover" the unobserved confounders for the Granger causality. We use a generative model with latent variable to build the relationship between the unobserved confounders and the observed variables(tested variable and the proxy variables). The posterior distribution of the latent variable is adopted to represent the confounders distribution, which can be sampled to get the estimated confounders. We adopt the variational autoencoder to estimate the intractable posterior distribution. The recurrent neural network is applied to build the temporal relationship in the data. We evaluate our method in the synthetic and semi-synthetic dataset. The result shows our estimated confounders has a better performance than the proxy variables in the non-linear Granger causality with multiple proxies in the semi-synthetic dataset. But the performances of the synthetic dataset and the different noise level of proxy seem terrible. Any advice can really help.
\end{abstract}

\section{Introduction}
\label{section:introduction}
Understanding the causal relationship between time series is a substantial problem in many domains. 
Examples include the causal relationship between stock prices and exchange rates in finance and the relationship between climate and vegetation in ecology. 
Granger causality(G-causality) test\cite{granger}, as one of the most popular time series causal analysis methods, has highlighted the its importance in many application studies\cite{alagidede_causal_2011,papagiannopoulouNonlinearGrangercausalityFramework2017,ozturkLongrunCausalAnalysis2013}. 


The most crucial aspect of inferring causal relationships from observational data is confounding. A variable which affects both the cause and the outcome is known as a \textit{confounder} of the effect of the cause on the outcome.
On the one hand, if such a confounder is observable, the standard way to account for its effect is by “controlling” for it, often through conditional Granger test\cite{liao_evaluating_2010}. 
On the the other hand, if a confounder is hidden or unobservable, it is impossible in the general case (i.e. without further assumptions) to estimate the effect of the cause on the outcome\cite{geigeraCausalInferenceIdentification,petersCausalInferenceTime}. 
For example, economic growth can affect both the tax price, and the consumption level of individual residents. Therefore financial development acts as confounder between tax price and consumption level of individual residents. Without measuring it we cannot in general isolate the cause effect of tax price on the consumption level of individual residents. 

In most real-world observational studies we cannot hope to measure or define all possible confounders. For example, it is still an open question how to quantify the financial development in finance\cite{RePEc:eee:chieco:v:17:y:2006:i:4:p:395-411}. Meanwhile, the unobservable confounders will lead to the spurious Granger causality problem\cite{maziarzReviewGrangercausalityFallacy2015}.

In practical research, people often rely on so-called "proxy variables"\cite{RePEc:eee:chieco:v:17:y:2006:i:4:p:395-411,ray_testing_2012}. For example, we cannot quantify the financial development, we might be able to get a proxy for it by using the a ratio of bank deposit liabilities to income\cite{RePEc:eee:chieco:v:17:y:2006:i:4:p:395-411}. Normally, the proxies are treated as noisy measurements of the confounders\cite{pearl_measurement_nodate,louizosCausalEffectInference}.
However, it's well known\cite{pearl_measurement_nodate,kuroki_measurement_nodate}that it is often incorrect to control the proxies' effect as if they are ordinary confounders, as this would induce bias. There are still the debates about the proxy variable in the practical work, because different proxy variables may lead to distinct conclusions\cite{KAR2011685}.

Therefore, \textbf{the main problem(challenge) is how to estimate the Granger causality under unobservable confounders, even with proxy variables}. We propose to estimate the unobservable confounders by sampling from its posterior distribution conditioning on the observable variables. It is a richer information for the unobservable confounders than just proxy in observational studies. Based on the causal assumption between the unobservable confounders and the observable variables, we construct a generative model with latent variable, then represent the posterior distribution of unobservable confounders with the posterior distribution of latent variable. To handle the intractable problem of the posterior distribution of latent variable in the generative model, we use variational autoencoders(VAEs) to approximate it. Through the causal relationship constraint in the generative model, the estimated confounder can be interpreted as the "denoised" proxy variable, because it is the common cause of the observable variables.

To model the posterior distribution of the unobservable confounders, we have to consider the temporal dependence and causal relationship of between the unobservable confounders and observable variables in the generative model, \textbf{This is the second challenge for us}. 
In this paper, we use the gated recurrent unit(GRU), a special kind of recurrent neural network(RNN) to model the time series, where RNN can exhibit temporal dynamic behavior. Specifically, the internal states in LSTM of the cause time series, are used to model its causal impact on the effect. We don't need restrict the time lag of the causal relationship, because the internal states contain the past information of the cause time series.

{\color{red}We evaluate our method on the both synthetic data and real data. However, we don't get the expected result. The estimated unobservable confounders cannot compete with the proxy variable in the performance of Granger test. So we expect your comments about our method.}

\begin{figure}[t]
\centering
\vspace{-0.8cm} 

\includegraphics[width=6cm]{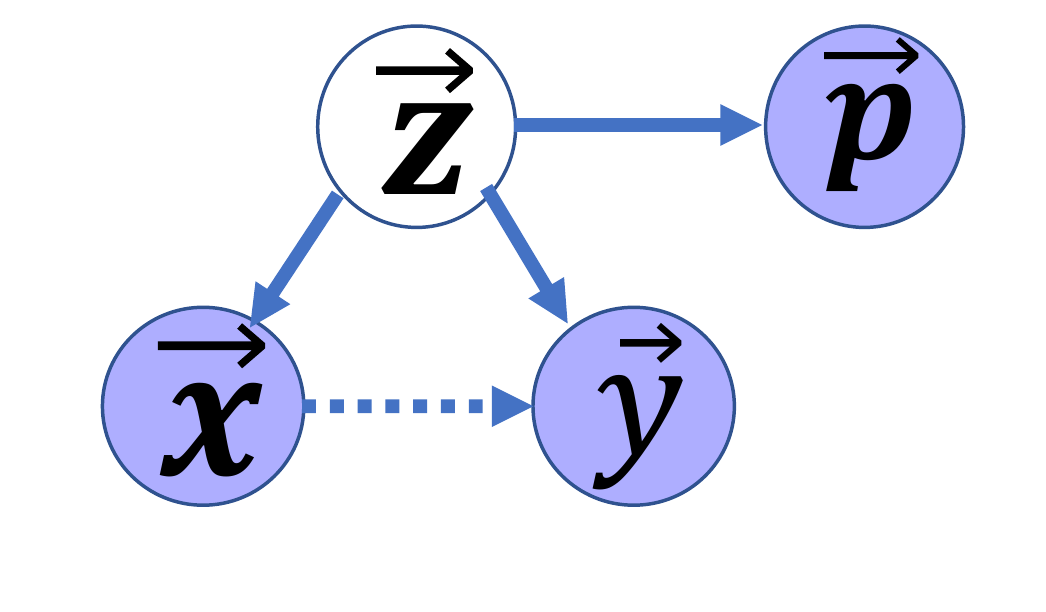}

\caption{Causal graph between tested time series $X$, $Y$, unobservable confounders $Z$, and proxy time series $P$ }
\label{fig:causal_graph}
\end{figure}

\section{Related Work}
\label{section:related} 
\subsection{Causal analysis via proxy variable} In the empirical time series causality studies, proxy variables usually are used as a substitute for the unobservable(or undefined) confounders. In the investigation of the causal relationship between military expenditure and economic growth\cite{chang_military_2011}, the investment/GDP ratio is used to  proxy the confounder capital stock. The prime lending rate has been used as a proxy for the rate of interest in the study of the granger causality from foreign direct investment to economic growth\cite{campbell_impact_2012}, where the rate of interest is a undefined confounder. However there are lots of the debates for the proxies choice in the practical problems\cite{vitasse_is_2014,hobbs_is_2010}. Because the proxy variables usually are affected by lots of features, which may induce the bias for the causal analysis, even the wrong results. There are studies \cite{pearl_measurement_nodate,kuroki_measurement_nodate} point potential bias in the causal analysis, when we directly control the proxies' effect as they are the ordinary confounders. But there are still no solutions for this problem in the time series studies as far as we know. The proxy problem in the static data causal analysis has been studied\cite{louizosCausalEffectInference}, where the cause is a binary variable, e.g., having the medicine or not, therefore they don't need consider the temporal relationship in the data.

\subsection{Causal analysis via latent variable model}
Unobservable variables are the fundamental problems in the observational causality study, such as, the unobservable confounders, the unknown state in the counterfactual problems. In recent years, several works try to handle these problems with the help of the latent variable model. \cite{louizosCausalEffectInference} use the latent variable model to learn the causal relationship with the unknown confounders. The estimated posterior distribution of latent variable, is used to learn the individual-level causal effects. Their study is based on the static data, where the cause is a binary variable. They use the bayesian neural network to model the causal relationship between variables and VAE to handle the intractable posterior distribution. \cite{krishnanDeepKalmanFilters2015} design a temporal generative model to learn the causal effect of the intervention, such as how one new treatment on a patient affects the blood sugar level. Their goal is to answer the counterfactual question, such as what the patient's blood level will be if he didn't take the pill. It is unknowable in the real studies, because one patient can only accept one treatment. In their work, they use a latent variable to represent the unmeasurable patient health state, which is effected by the interventions and impacts the observable patient record data. They take the RNN to model the temporal relationship and VAE to train the model. After training the model with the observed data, they sample on the latent state distribution to get expected effect of different interventions. In their work, the latent variable is the cause of the patient records and the outcome of the intervention in the mean time, which is not coincident with our assumption of the problem.

\section{Background}
\label{section:background} 
\subsection{Granger Causality}
Granger Causality, which was provided by C.W.J Granger in 1969, has been used in economics\cite{alagidede_causal_2011}, neuroscience\cite{seth_granger_2015}, climatology\cite{papagiannopoulouNonlinearGrangercausalityFramework2017}, etc., in recent years. The fundamental assumption of the Granger causality is \cite{berzuini_causal_2012}:
\begin{itemize}
\item The effect does not precede its cause in time.
\item The causal series contains unique information about the series being caused
that is not available otherwise.
\end{itemize}
The assumption two is often justified through the prediction model in practice. Here we denote the time series $\vec{x} = (x_1,x_2,\cdots,x_T)$ and  time series $\vec{y} = (y_1,y_2,\cdots,y_T)$.
If the accuracy of prediction $y_t$ will be improved significantly by involving $x_{t-1},\cdots, x_{t-\mu}$,	 compared to considering the $y_{t-1},\cdots, y_{t-\tau}$ only, where $\mu$ is the max time lag for $\vec{x}$, $\tau$ is the max time lag for $\vec{y}$, we can say $\vec{x}$ is the Granger-cause of $\vec{y}$. 
Here we define the Granger test from $\vec{x}$ to $\vec{y}$ as $GC(\vec{x} \to \vec{y})$. If there are observable confounders, such as $\vec{\textbf{m}}$, $\textbf{m}_t\in\mathbb{R}^p$, we can take the conditional Granger test, denoted as $GC(\vec{x} \to \vec{y}|\vec{\textbf{m}})$. The effect of the common cause will be eliminated by adding $\textbf{m}_{t-1},\cdots, \textbf{m}_{t-\sigma}$ in the both prediction model.

As a crucial problem in observational studies, unobservable confounder may lead to the spurious Granger causality problem, here we use an example to explain it\cite{maziarzReviewGrangercausalityFallacy2015}. Assume $\vec{z}$ is the unobservable confounder, $z_t$ causes $x_{t+2}$ and $y_{t+4}$. If we test whether $\vec{x}$ Granger-cause $\vec{y}$, the $x_{t-\tau}$ may well improve the precision of the prediction of $y_t$, because $z_{t-4}$ is the common cause of $y_t$ and $x_{t-2}$, which is not be controlled. 
\subsection{Variational Autoencoder}
Deep Bayesian networks use neural networks to express the relationships between variables, which often leads the intractability of posterior inference. In recent years, \cite{rezende_stochastic_2014,kingmaAutoEncodingVariationalBayes2013} introduce the variational inference techniques to handle this problem, which is VAE. The key technique of VAE is the inference network, a neural network which approximates the intractable posterior. Via the reparameterization trick, the inference network can be trained with the generative network together.

Let's consider the generative model of the observable variable $x$, $p(x)=\int p_{\theta}(x|z)p_0(z) dz$, where the $p_0(z)$ is the prior distribution of latent variable $z$ and $p_{\theta}(x|z)$ is a generative model parameterized by $\theta$. The true posterior distribution $p_{\theta}(z|x)$ is typically intractable, when the $p_{\theta}(x|z)$ is too complex, such as a neural network. VAE uses the variational inference techniques to fit the approximate posterior distribution $q_{\phi}(z|x)$ by neural network, named inference network. Typically, VAE takes SGVB, a variational inference algorithm, to jointly train the approximated posterior and the generative model by maximizing the evidence lower bound(ELBO, Eq \ref{eq:elbo}).
\begin{equation}
\begin{aligned}
log\ p_{\theta}(x) &\ge  log\  p_{\theta}(x)-KL[q_{\phi}(z|x) \| p_{\theta}(z|x)] \\
&= \mathcal{L}(x;(\theta,\phi)) \\
&= \mathbb{E}_{q_{\phi}(z|x)}[log\  p_{\theta}(x) + log\ p_{\theta}(z|x)-log\ q_{\phi}(z|x)] \\
&= \mathbb{E}_{q_{\phi}(z|x)}[log\  p_{\theta}(x, z)-log\ q_{\phi}(z|x)] \\
&= \mathbb{E}_{q_{\phi}(z|x)}[log p_{\theta}(x|z) + log\ p_{\theta}(z)-log\ q_{\phi}(z|x)]
\end{aligned}
\label{eq:elbo}
\end{equation}

The challenge for the optimization problem is the expectation on $q_{\phi}$ in the ELBO, which implicitly depends on the network parameters $\phi$.
Monte Carlo method can be used to estimate the expectation as Eq \ref{eq:monte}, when the latent state follow the specific distributions, such as normal distribution.

\begin{equation}
\mathbb{E}_{q_{\phi}(z|x)}[f(z)] \approx \frac{1}{L}\sum_{l=1}^{L}f(z^{(l)})
\label{eq:monte}
\end{equation}
$z(l),l =1 \cdots L$ are samples from $q_{\phi}(z|x)$.

\subsection{Recurrent neural network}
Recurrent neural network(RNN) is used to model the time series or sequence $\vec{\textbf{x}}$ by recursively processing each timestep value while maintaining its internal internal state $\textbf{h}$. At each timestep $t$, 
the RNN read $\textbf{x}_t \in \mathbb{R}^d$ and updates its internal state $\textbf{h}_t \in \mathbb{R}^p$
 by $\textbf{h}_t=f_{\theta}(\textbf{x}_t,\textbf{h}_{t-1})$, 
 where $f$ is a deterministic non-linear transition function and $\theta$ is the parameter of $f$.
The transition function $f$ can be implemented with gated activation functions such as long short-term memory(LSTM)\cite{doi:10.1162/neco.1997.9.8.1735} or gated recurrent unit(GRU) \cite{cho_learning_2014}. RNN can model time series by parameterizing a factorization of the joint time series probability distribution as a product of conditional probabilities as: 
\begin{equation}
\begin{aligned}
p(\textbf{x}_1,\textbf{x}_2,\cdots,\textbf{x}_T)&=p(\textbf{x}_1) \prod_{t=2}^Tp(\textbf{x}_t|\textbf{x}_1,\cdots,\textbf{x}_{t-1})
\\ 
p(\textbf{x}_t|\textbf{x}_1,\cdots,\textbf{x}_{t-1})&=g(\textbf{h}_{t-1})
\end{aligned}
\end{equation}
where $g$ is a function to map the RNN internal state $\textbf{h}_{t-1}$ to a probability distribution over possible outputs.


%
%

%
\section{Granger Causality via Variational Autoencoder}
\label{section:methodology} 
\textbf{Problem statement}:
Here we define the causal relationship we investigate in this paper as figure \ref{fig:causal_graph}. Our final goal is to estimate the G-causality from $\vec{x}$ to $\vec{y}$.
We assume a sufficient set $\vec{\textbf{z}}$ of confounders is unobservable. The observable proxy time series $\vec{\textbf{p}}$ , a measurement of $\vec{\textbf{z}}$ with noise.
are observable. The $\vec{x}$ , $\vec{y}$ are one-dimensional time series, $\vec{\textbf{p}}$ can be multi-dimensional time series and $\vec{\textbf{z}}$ is not known for us.
\subsection{Architecture}
Based on the causal graph in figure \ref{fig:causal_graph}, we assume the ground truth of G-causality from  $\vec{x}$ to $\vec{y}$ is $GC(\vec{x} \to \vec{y}|\vec{\textbf{z}})$. 
%
%
Comparing to the traditional methods, who use a proxy times series $\vec{\textbf{p}}$ to represent $\vec{\textbf{z}}$, we adopt the sampled  $\vec{\textbf{z}}$ from the estimated posterior distribution $p(\vec{\textbf{z}}|\vec{x},\vec{y},\vec{\textbf{p}})$ based on two intuitions:
\begin{itemize}
\item The posterior distribution $p(\vec{\textbf{z}}|\vec{x},\vec{y},\vec{\textbf{p}})$, includes the richest information about $\vec{\textbf{z}}$ based on the observable dataset. It can be more helpful in the Granger test.
\item Through the restriction of the learning process of the posterior distribution, we can treat it as the "denoising" process for $\vec{\textbf{p}}$. Because it may hold the noise information which is not the common cause for $\vec{x},\vec{y}$, then leads the bias in granger test.
\end{itemize}

\begin{figure}
\centering
\includegraphics[width=\linewidth]{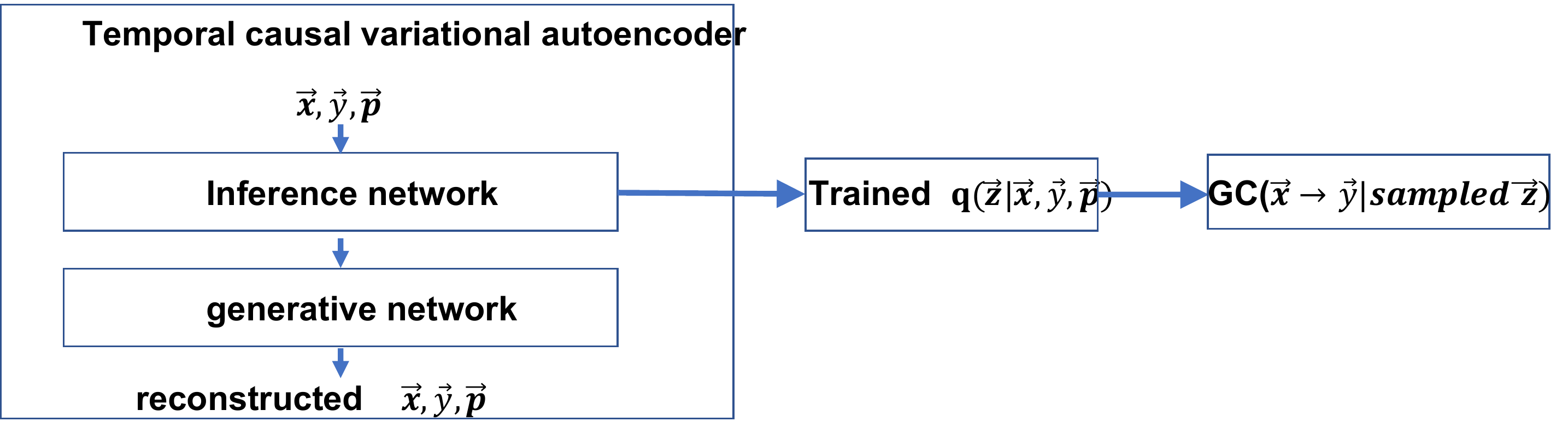} 
\caption{The overall architecture of \dname{}}
\label{fig:all}
\end{figure}

For the real dataset, it is not rare for the existence of the non-linear causal relationship\cite{papagiannopoulouNonlinearGrangercausalityFramework2017}. We assume the non-linear causal relationship in this paper and it will be modeled through neural network, which makes the true posterior distribution intractable. Therefore, we build a generative network for the observable time series. We can get the approximate posterior distribution $q(\vec{\textbf{z}}|\vec{x},\vec{y},\vec{\textbf{p}})$ through inference network of VAE. We use the sampled $\vec{\textbf{z}}$ from the trained approximate posterior distribution to represent the true $\vec{\textbf{z}}$. It can be adapted to kinds of Granger test methods. The whole architecture of our method \dname{} is shown in figure \ref{fig:all}.

\subsection{Temporal causal variational autoencoder}
\begin{figure*}[]
    \centering
    \begin{subfigure}[b]{0.45\textwidth}
    \centering
        \includegraphics[width=\linewidth]{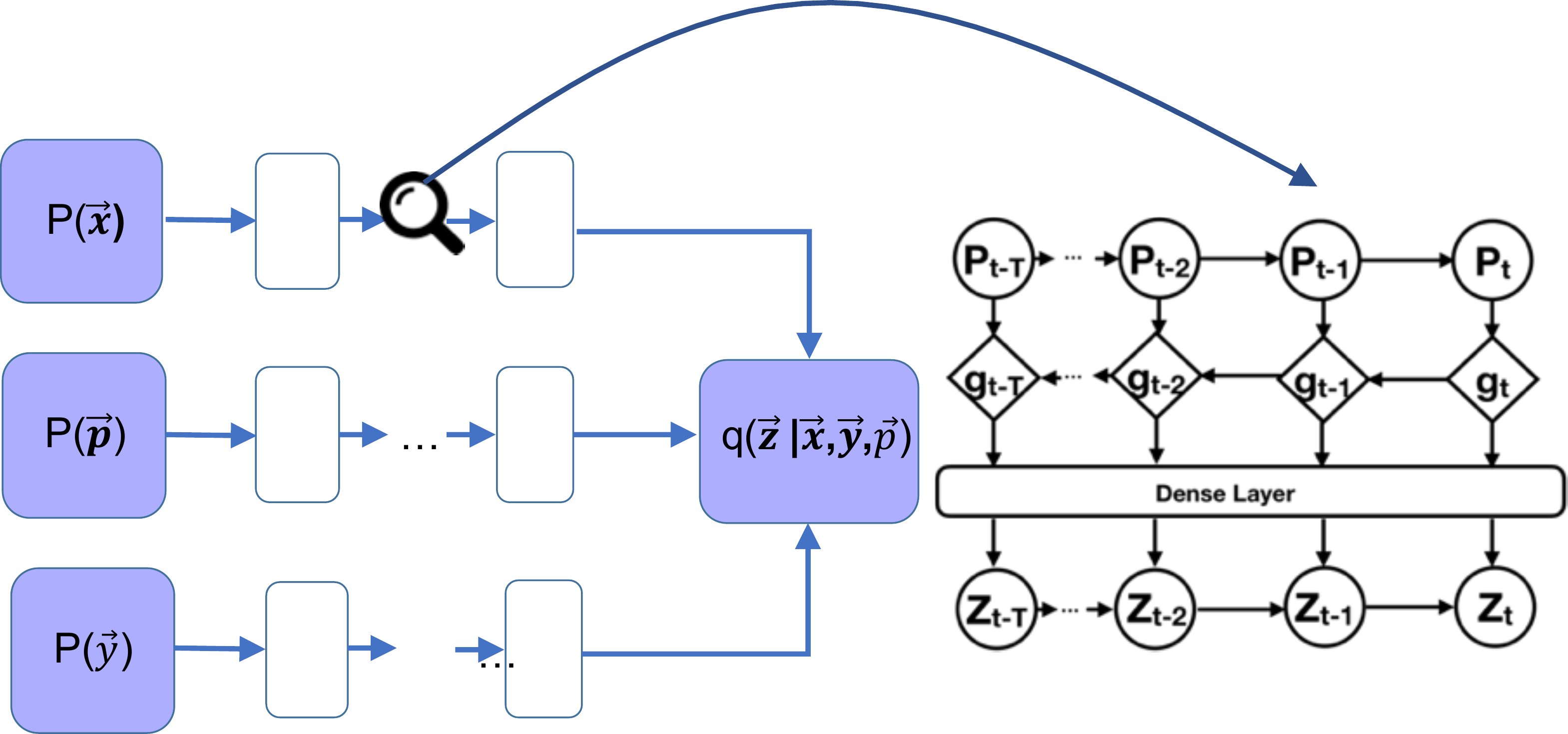}         
         \caption{Inference network} 
         \label{fig:inference}
    \end{subfigure}
    \begin{subfigure}[b]{0.45\textwidth}  
    \centering
        \includegraphics[width=\linewidth]{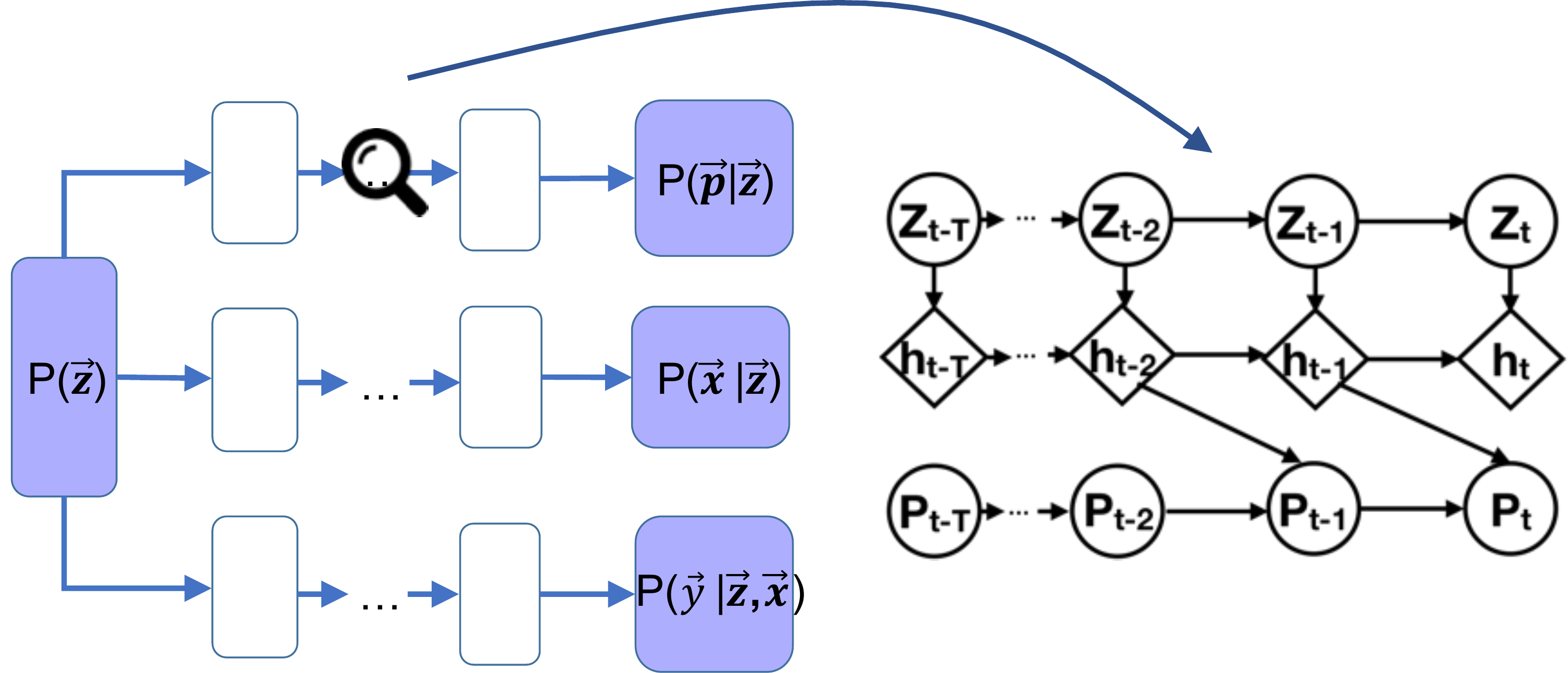} 
        \caption{Generative network} \label{fig:generative}
    \end{subfigure}
       \caption{The overall architecture of the temporal causal variational autoencoder. For the left part of each subfigure, the blank nodes correspond to parametrized deterministic neural network, filled nodes correspond to drawing samples from the respective distribution.}
    \label{fig:vae}
\end{figure*}

The overall architecture of the generative network and inference network for the temporal causal variational autoencoder is shown in figure \ref{fig:vae}. 
\textbf{Generation}: Each generative path in the generative network, e.g., $\vec{\textbf{z}}$ to $\vec{\textbf{p}}$, also embraces the time series causal relationship. The difficulty to model this relationship is the time lag. The $\textbf{p}_t$ may not just be caused by the $\textbf{z}_t$. All the causal relationship between $\textbf{z}_{t-\tau}$ and  $\textbf{p}_t$ is possible.
Unlike the past RNN generative model, e.g., the Deep Kalman Filters, we choose the internal state $\textbf{h}_{t-1}$ in GRU of $_{t-1}$ to generate $\textbf{p}_{t}$. The $\textbf{h}_{t-1}$ as the summary of the past information of $\textbf{z}_t$, represents the causes from the past $\textbf{z}_t$, which releases us from the restriction of the causal time lag. Here we can choose $\textbf{h}_t$ or $\textbf{h}_{t-1}$ to generate $\textbf{p}_t$, which builds the instantaneous or noninstantaneous causal relationship. The temporal relationship in the generative model is shown in figure \ref{fig:generative}. 

The prior $p_\theta(\vec{\textbf{z}})$ follows the linear Gaussian State Space Model\cite{Kitagawa1996} to keep the temporal relationship in the latent space:$z_t=O_{\theta}(T_{\theta}z_{t-1}+v_t)+\epsilon_t$, where $(T_{\theta}$ and $O_{\theta}$ are transition and observation matrices, $v_t$ and $\epsilon_t$ are transition and observation noises. For each time point of $\vec{\textbf{p}}$(or $\vec{\textbf{x}}$), the generating distribution will be conditioned on both $\textbf{p}_{t-1}$(or $\textbf{x}_{t-1}$) and the internal state of the GRU $\textbf{h}_{t-1}$, which is:

\begin{equation}
\begin{aligned}
p_{\theta}(x_t|x_{t-1},\vec{\textbf{z}}) &= \mathcal{N}(\mu_{x_t},\sigma^2_{x_t})\\
p_{\theta}(\textbf{p}_t|\textbf{p}_{t-1},\vec{\textbf{z}}) &= \mathcal{N}(\pmb{\mu}_{\textbf{p}_t},diag(\pmb{\sigma}^2_{\textbf{p}_t}))
\end{aligned}
\end{equation}
where $[\mu_{x_t},\sigma^2_{x_t}]=\varphi^{x_t}_\theta(x_{t-1}, \textbf{h}_{t-1}^z)$,  
$[\pmb{\mu}_{\textbf{p}_t},\pmb{\sigma}^2_{\textbf{p}_t}]=\varphi_\theta^{\textbf{p}_t}(\textbf{p}_{t-1}, \textbf{h}_{t-1}^\textbf{z})$. 
$\varphi_\theta^{\textbf{p}_t}$ and $\varphi^{x_t}_\theta$ are the neural networks in the generative network parameterized by $\theta$. $\textbf{h}_{t-1}^\textbf{z}$ is the internal state of GRU for $\vec{\textbf{z}}$, which is updated by the recurrence equation:$\textbf{h}_{t}^\textbf{z}=f_{\theta}(\textbf{z}_t,\textbf{h}_{t-1}^\textbf{z})$, where $f_\theta$ is the transition function of GRU. Therefore $\textbf{h}_t^\textbf{z}$ is a summary of the past information of $\textbf{z}_t$ and the current value of $\textbf{z}_t$. The generation distribution of $y_t$ is $p_{\theta}(y_t|y_{t-1},\vec{\textbf{z}})= \mathcal{N}(\mu_{y_t},\sigma^2_{y_t})$, where $[\mu_{y_t},\sigma^2_{y_t}]=\varphi^{y_t}_\theta(y_{t-1}, \textbf{h}_{t-1}^\textbf{z}, \textbf{h}_{t-1}^x)$ to construct the causal relationship between $\vec{x}$ and $\vec{y}$. 

%
%
\textbf{Inference}: Based on the generative model(figure \ref{fig:generative}), we note the true posterior distribution is factorized as:

\begin{equation}
\begin{aligned}
p(\vec{\textbf{z}}|\vec{x},\vec{y},\vec{\textbf{p}}) = p(\textbf{z}_1|\vec{x},\vec{y},\vec{\textbf{p}})\prod_{t=2}^Tp(\textbf{z}_t|\textbf{z}_{t-1},\vec{x},\vec{y},\vec{\textbf{p}})
\end{aligned}
\end{equation}
By the Markov structure of our generative graphical model we notice $\textbf{z}_t \perp (x_1,\cdots,x_{t-1}) | \textbf{z}_{t-1}$,  $\textbf{z}_t \perp (y_1,\cdots,y_{t-1})| \textbf{z}_{t-1}$ and $\textbf{z}_t \perp (\textbf{p}_1,\cdots,\textbf{p}_{t-1})| \textbf{z}_{t-1}$, therefore we get:
\begin{equation}
\begin{aligned}
&p(\vec{\textbf{z}}|\vec{x},\vec{y},\vec{\textbf{p}})\\
&= p(\textbf{z}_1|\vec{x},\vec{y},\vec{\textbf{p}})\prod_{t=2}^Tp(\textbf{z}_t|\textbf{z}_{t-1},\vec{x},\vec{y},\vec{\textbf{p}}) \\
&= p(\textbf{z}_1|\vec{x},\vec{y},\vec{\textbf{p}})\prod_{t=2}^Tp(\textbf{z}_t|\textbf{z}_{t-1},x_t,\cdots,x_T,y_t,\cdots,y_T,\textbf{p}_t,\cdots,\textbf{p}_T) 
\end{aligned}
\end{equation}

Therefore, we can use the reverse-RNN\cite{krishnanDeepKalmanFilters2015} to design the inference network to approximate the true posterior distribution. The details of the inference network is shown in \ref{fig:inference}. For each observed time series, a reverse GRU is modeled, which is:
\begin{equation}
\begin{aligned}
\textbf{g}_{t-1}^\textbf{p} &= f_{\phi}(\textbf{g}_t^\textbf{p}, \textbf{p}_{t-1})\\
\textbf{g}_{t-1}^x &= f_{\phi}(\textbf{g}_t^x, x_{t-1})\\
\textbf{g}_{t-1}^y &= f_{\phi}(\textbf{g}_t^y, y_{t-1})\\
\end{aligned}
\end{equation}
where $f_{\phi}$ is the transition function of GRU parameterized by $\phi$. Therefore $\textbf{g}_{t}^y$ is the summary of the information of $y_t,\cdots,y_T$, so as $\textbf{g}_{t}^x, \textbf{g}_{t}^\textbf{p}$. Therefore the approximate posterior distribution for $z_t$ is:
\begin{equation}
\begin{aligned}
q(\textbf{z}_t|\textbf{z}_{t-1},x_t,\cdots,x_T,y_t,\cdots,y_T,\textbf{p}_t,\cdots,\textbf{p}_T) \\
= \mathcal{N}(\pmb{\mu}_{\textbf{z}_t},diag(\pmb{\sigma}_{\textbf{z}_t}^2))\\
where [\pmb{\mu}_{\textbf{z}_t},\pmb{\sigma}^2_{\textbf{z}_t}] = \varphi_{\phi}^{\textbf{z}_t}(\textbf{z}_{t-1},\varphi_{\phi}(\textbf{g}_{t}^x, \textbf{g}_{t}^\textbf{p}, \textbf{g}_{t}^y))
\end{aligned}
\end{equation}
where $\varphi_{\phi}^{\textbf{z}_t}$,$\varphi_{\phi}$ is the neural network in the inference network.

\begin{figure}
\centering
\includegraphics[width=\linewidth]{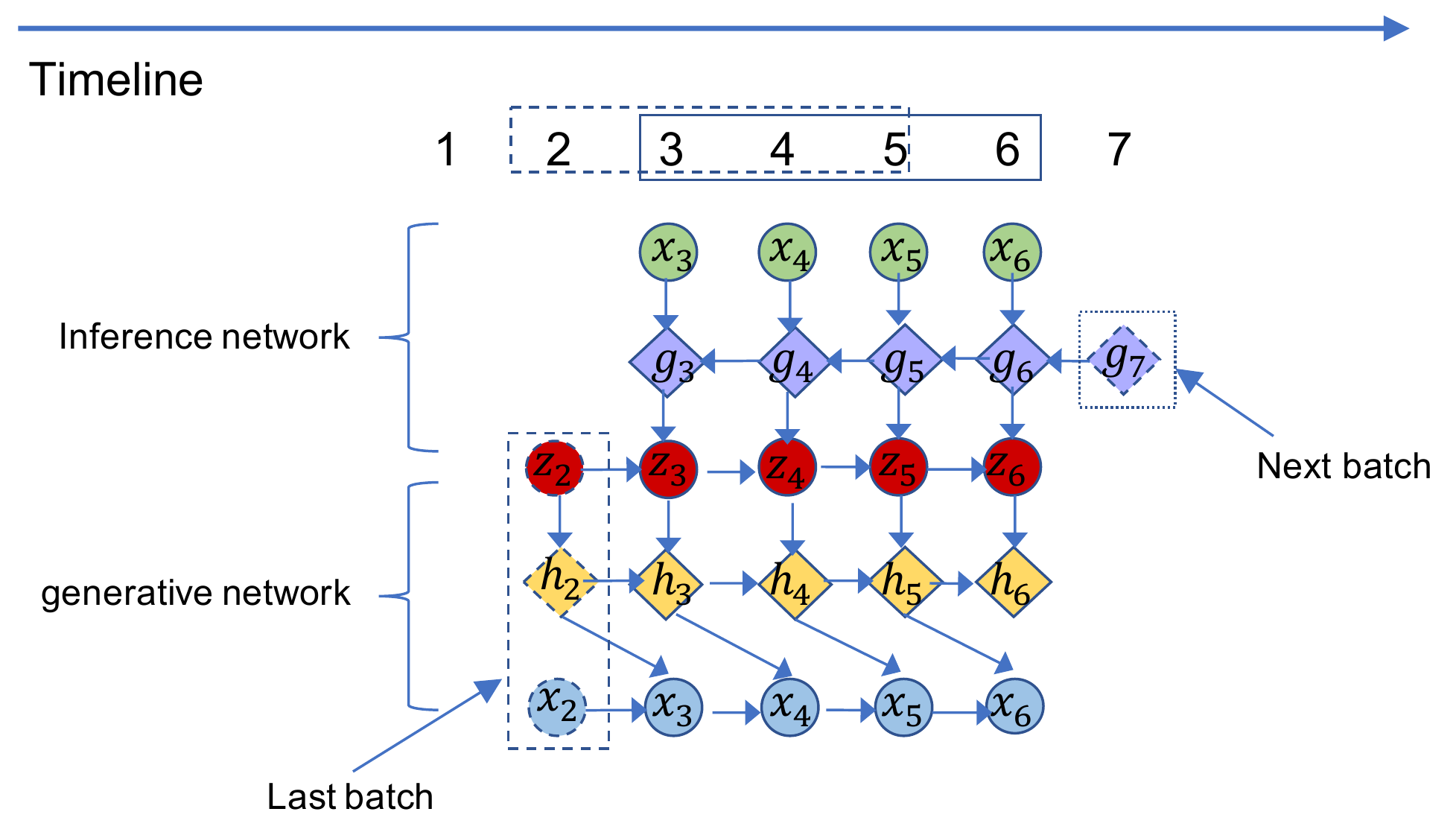} 
\caption{The example for the sliding window training, where the length $L$ of the sliding window is 4.}
\label{fig:slide_window}
\end{figure}

\textbf{Training}: In practice, there is only one sample for $\vec{x}$ or $\vec{y}$. Granger test will set the max time lag to construct the several time slice samples for the  statistical examination. Mirroring it, we use a sliding window to conduct the batch sample for the neural network training, where the batch size is 1 (shown in figure \ref{fig:slide_window}). 
One sliding window whose length is $L$, contains the data $x_i,\cdots,x_{i+L-1},y_i,\cdots,y_{i+L-1},\textbf{p}_i,\cdots,\textbf{p}_{i+L-1}$, where $i$ is the index for the batch. The window will slide one time point each time. 
To keep the temporal relationship between the two adjacent batches, in the generative network, the $\textbf{h}_i^\textbf{z}$ of batch $i$, will be used to initialize the RNN internal state in batch $i+1$. Considering the reverse RNN in the inference network, we will use the $q(\textbf{z}_i|\textbf{z}_{i-1},x_i,\cdots,x_{i+L-1},y_i,\cdots,y_{i+L-1},\textbf{p}_i,\cdots,\textbf{p}_{i+L-1}$ in batch $i$ to sample the estimated $\textbf{z}_i$, we assume it approximates $q(\textbf{z}_i|\textbf{z}_{i-1},x_i,\cdots,x_{T},y_i,\cdots,y_{T},\textbf{p}_i,\cdots,\textbf{p}_{T})$.
According to the variational inference techniques, the parameters in the generative network $\theta$ and the inference network $\phi$ can be optimized simultaneously by maximizing the variational ELBO in \label{section:background}:

\begin{equation}
\begin{aligned}
&\mathcal{L}(\vec{x},\vec{y},\vec{\textbf{p}}) \\
&= \mathbb{E}_{q_{\phi}(\vec{\textbf{z}}|\vec{x},\vec{y},\vec{\textbf{p}})}[log\ p_\theta(\vec{y}|\vec{x},\vec{\textbf{z}})+log\ p_\theta(\vec{x}|\vec{\textbf{z}})\\
&\qquad\qquad\qquad+log\ p_\theta(\vec{\textbf{p}}|\vec{\textbf{z}}) + log\ p_\theta(\vec{\textbf{p}})] 
\end{aligned}
\end{equation}
The trained $q_{\phi}(\vec{\textbf{z}}|\vec{x},\vec{y},\vec{\textbf{p}})$ will be used to sample the estimated $\vec{\textbf{z}}$ for the Granger test.


\section{Experiment}
\label{section:exp}
Evaluating causal analysis method is always challenging because we usually lack ground-truth of causal effect. Common evaluation approaches include creating synthetic or semi-synthetic datasets, where real data is modified in a way that allows us to know the true causal effect.
Here We conduct a qualitative analysis of synthetic datasets, with the ground-truth is the causality exists or not.
With the semi-synthetic and real dataset, we conduct the causal analysis with the coefficient of determination $R^2$, which can be used to define the Granger causality quantitatively\cite{papagiannopoulouNonlinearGrangercausalityFramework2017}. 
For all kinds of dataset, we compare the performance of our method \textbf{\dname{}} and the baseline method \textbf{Granger} based on non-linear model. 
\subsection{Experiment on synthetic data}
To illustrate that our method handles hidden confounders better we experiment on a toy simulated dataset who can be controlled whether the causality exits.
%
%

First we conduct the qualitative analysis of causality.
The non-linear synthetic dataset is generated by the following process:
\begin{equation}
\begin{aligned}
		z_t &= tanh(z_{t-1}) + \mathcal{N}(0, 1)\\
		x_t &= tanh(\frac{2}{3} * z_{t - 2} + \frac{1}{3} * z_{t - 1})\\ &+ \frac{\frac{1}{3} * x_{t - 2} + \frac{2}{3} * x_{t - 1}}{4} + 0.05 \mathcal{N}(0, 1) \\
		y_t &= sigmoid(\frac{1}{3} * z_{t - 4} + \frac{2}{3} * z_{t - 3})\\ 
		&+ \frac{\frac{1}{3} * y_{t - 2} + \frac{2}{3} * y_{t - 1}}{4} + 0.05 \mathcal{N}(0, 1) \\
		p(t) &= z_t + ploss * \overline{z} * \mathcal{N}(0.5, 1)	
\end{aligned}
\end{equation}
where $\overline{z}$ is the mean of the $\vec{z}$.


We generate 500 datasets of all the time series(length 1000). Therefore, we don't need to adopt the sliding window to conduct the training samples.

\begin{figure}[h]
    \centering
    \begin{subfigure}[b]{0.23\textwidth}
    \centering
        \includegraphics[width=\linewidth]{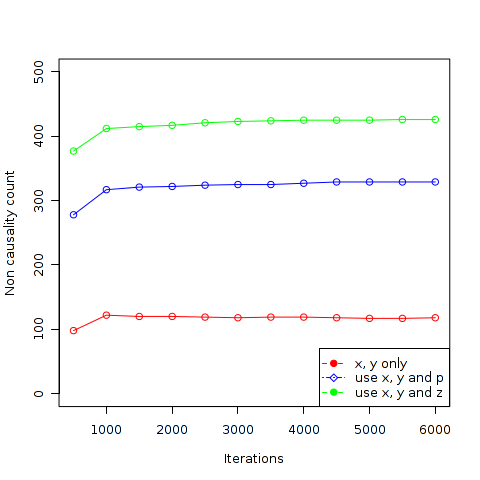}         
         \caption{$ploss=1, D_{\vec{\textbf{p}}}=2$} 
         \label{fig:inference}
    \end{subfigure}
    \begin{subfigure}[b]{0.23\textwidth}  
    \centering
        \includegraphics[width=\linewidth]{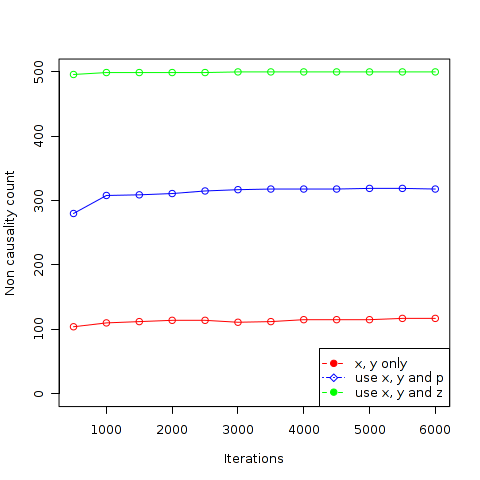} 
        \caption{$ploss=1, D_{\vec{\textbf{p}}}=5$} 
        \label{fig:generative}
    \end{subfigure}
    \begin{subfigure}[b]{0.23\textwidth}
    \centering
        \includegraphics[width=\linewidth]{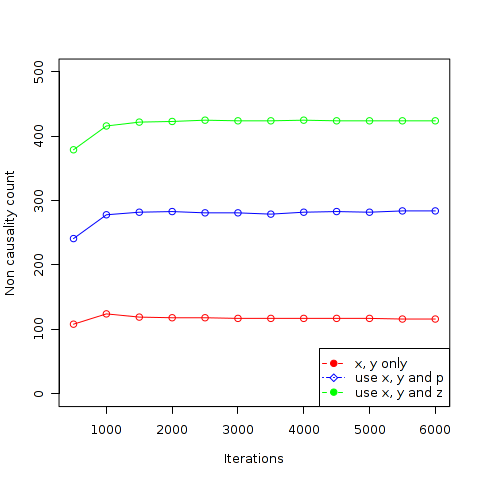}         
         \caption{$ploss=3, D_{\vec{\textbf{p}}}=2$} 
         \label{fig:inference}
    \end{subfigure}
    \begin{subfigure}[b]{0.23\textwidth}  
    \centering
        \includegraphics[width=\linewidth]{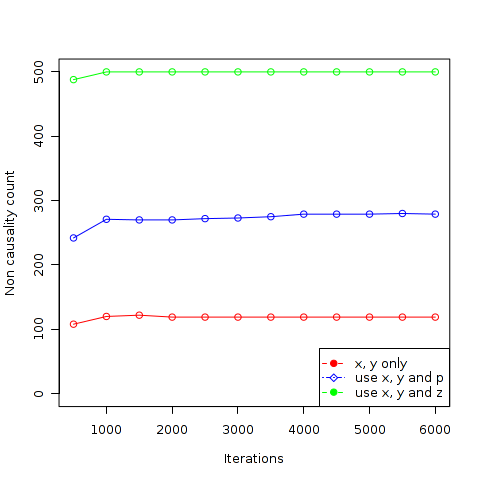} 
        \caption{$ploss=3, D_{\vec{\textbf{p}}}=5$} 
        \label{fig:generative}
    \end{subfigure}
    \caption{The result of synthetic data with qualitative analysis of causality. The x axis is the number of iterations for the non-linear Granger test in NlinTS, The y axis is the count of the false result.}
       
    \label{fig:simulation_xy_not}
\end{figure}

Based on the generative process, we know the ground truth is there is no causal relationship between $\vec{x}$ and $\vec{y}$. So we compare the result of non-linear Granger test conditioning on the $\vec{\textbf{p}}$ and the estimated confounder $\vec{\textbf{z}}$. The method who can judge the causal relationship is better. The results is shown in figure \ref{fig:simulation_xy_not}.

\begin{figure}[h]
    \centering   
        \includegraphics[width=0.8\linewidth]{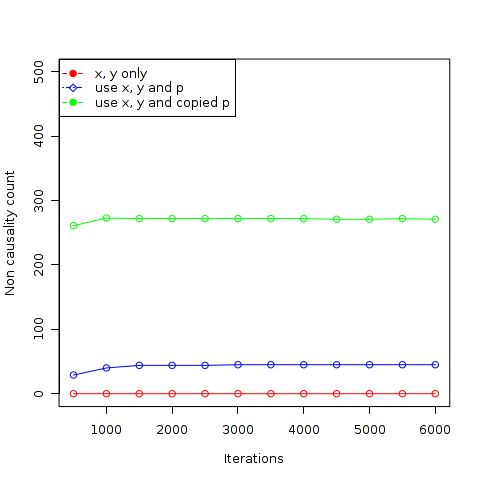}         
         \caption{The result of the non-linear Granger causality with different dimensionality of $\vec{\textbf{p}}$, where the "copied p" is the 5-d times series, and each dimension of  the "copied p" is the $\vec{p}$} 
         \label{fig:same_p}  
\end{figure}

{\color{red} We encounter one difficulty when we choose the non-linear conditional Granger test method. We didn't find any open source of the related algorithm. We modify the non-linear Granger test method in the R-package NlinTS\cite{NlinTS}. Following the classical Granger test, it uses the feed-forward neural networks to build the regression model, and applies the F-test for the Granger test. We add the conditional variables in the both regression models with and without $\vec{x}$. F-test is still used to do the Granger test. But we find this method is sensitive to the dimensionality of the conditional variables. Along with the growth of the dimensionality of the conditional variables, it tends to get the non causal conclusion, no matter the ground truth(shown in figure \ref{fig:same_p}). Because we can't solve this problem, we can't adjust the dimensionality of our estimated confounders. }

Then we conduct the quantitative analysis of the causality. The non-linear synthetic dataset is generated by the following process, where the $\vec{x}$ and $\vec{y}$ have the non-linear causal relationship:
\begin{small}
\begin{equation}
\begin{aligned}
		z_t &= tanh(z_{t – 1}) + \mathcal{N}(0, 1)\\
		w_t &= log(w_{t – 1} + 1) + \mathcal{N}(0, 1)\\
		x_t &= tanh(\frac{2}{3} * z_{t - 2} + \frac{1}{3} * z_{t - 1} - 1) \\
	&+ sigmoid(\frac{1}{10} * w_{t - 4} + \frac{2}{10} * w_{t - 3} + \frac{3}{10} * w_{t - 2} + \frac{4}{10} * w_{t - 1}) \\
	&+ \frac{\frac{1}{3} * x_{t - 2} + \frac{2}{3} * x_{t - 1}}{4} + 0.05 \mathcal{N}(0, 1)\\
	y_t &= sigmoid(\frac{1}{3} * z_{t - 4} + \frac{2}{3} * z_{t - 3}) \\
	&+ tanh(\frac{1}{3} * x_{t - 2} + \frac{2}{3} * x_{t - 1} - 1) \\
&+ \frac{\frac{1}{3} * y_{t - 2} + \frac{2}{3} * y_{t - 1}}{4}+ 0.05 \mathcal{N}(0, 1)\\
p_t& = z_t + ploss * \overline{z} * \mathcal{N}(0.5, 1)
\end{aligned}
\end{equation}
\end{small}

we use the coefficient of determination $R^2$ to evaluate our methods quantitatively\cite{papagiannopoulouNonlinearGrangercausalityFramework2017}. It is defined as follows:

\begin{equation}
\begin{aligned}
	R^2(\vec{y},\hat{\vec{y}}) = 1-\frac{RSS}{TSS}= 1 - \frac{\sum_{t=L+1}^N(y_t-\hat{y}_t)^2}{\sum_{t=L+1}^N(y_t-\bar{y})^2}
\end{aligned}
\end{equation}

where $\vec{y}$ represents the observed time series, $\bar{y}$ is the mean of model, $\hat{\vec{y}}$ is the predicted time series obtained from a given forecasting model, and $L$ is the length of the lag-time moving window. Therefore, the $R^2$ can be interpreted as the fraction of explained variance by the forecasting model, and it increases when the performance of the model increases. Therefore, we can define the quantitative Granger causality from time series $\vec{x}$ to $\vec{y}$ as the improvement of $R^2(\vec{y},\hat{\vec{y}})$, when the $x_{t-1},x_{t-2},\dots, x_{t-L}$ are included in the prediction of $y_t$, in contrast to considering $y_{t-1},y_{t-2},\dots, y_{t-L}$ only\cite{papagiannopoulouNonlinearGrangercausalityFramework2017}. For conditional $R^2$ improvement, the conditional variables will be included in the both predictive models. Following \cite{papagiannopoulouNonlinearGrangercausalityFramework2017}, we use random forest as the non-linear regression model.

The ground truth Granger causality is the improvement of $R^2$ conditioning on the $\vec{z}$. 
The performance of the Granger test method conditioning on $\vec{m}$ is $Diff(\vec{m})=|GC(\vec{x}\to\vec{y}|\vec{m})-GC(\vec{x}\to\vec{y}|\vec{z})|$.

Our result is shown in table \ref{tab:simulation_xy_exist}
\begin{table}[!hbtp]
\caption{	The experiment results of the synthetic data, where the $\vec{x}$ and $\vec{y}$ have the non-linear causal relationship.}
\vspace*{2mm} 
\label{tab:simulation_xy_exist}
\begin{center}
\begin{tabular}{|c|c|c|c|c|}
\hline
$ploss$ & $D_{\vec{\textbf{z}}}$ &  Diff($\vec{p}$) & Diff($\hat{\vec{z}}$) & Diff($\hat{\vec{z}}$)-Diff($\vec{p}$) \\
\hline
\hline
1 & 2 & 0.0046 & 0.0143  & 0.0097  \\
2 & 2 & 0.0075 & 0.0103 &  0.0027 \\
3 & 2 & 0.0082 & 0.0116 &  0.0034 \\
4 & 2 & 0.0071 & 0.0153 &  0.0082 \\
1 & 1 & 0.0078 & 0.0115 &  0.0036 \\
2 & 1 & 0.0075 & 0.0103 & 0.0028  \\
3 & 1 & 0.0068 & 0.0085 & 0.0018  \\
4 & 1 & 0.0069 & 0.0108 & 0.0039  \\
\hline
\end{tabular}
\end{center}
\vspace{-1cm}
\end{table}

%

\subsection{Experiment on semi-synthetic data}
To examine the performance of our method in the practical problem, we introduce two real datasets Butter\cite{Butter} and Temperature\cite{temperature}. The Butter dataset includes the weekly prices of the butter, milk and cheese in the U.S. from 2000 to now. We suppose the milk prices is the confounder for the causality relationship of milk price and cheese price\cite{petersCausalInferenceTime}. But the causality relationship is unclear. In our experiment, the $\vec{x}$ is the butter price, the $\vec{y}$ is the cheese price and the $\vec{z}$ is the milk price.
Temperature dataset is collected from a monitor system mounted in a domotic house. The data is sampled every minute, and smoothed with 15 minute means, spanning approximately 40 days. We choose the indoor temperature as $\vec{x}$, indoor humidity as $\vec{y}$ and outdoor temperature as $\vec{z}$, which is the confounder. Similar to \cite{louizosCausalEffectInference}, we construct the $\vec{p}$ by add some noise on the confounder, $p_t = z_t + N(0, \sigma^2)$, where $\sigma = noise \times mean(\vec{z})$. Therefore, we can evaluate the algorithm under different noise level. we also use the $Diff(\vec{m})=|GC(\vec{x}\to\vec{y}|\vec{m})-GC(\vec{x}\to\vec{y}|\vec{z})|$ as the indicator of the performance of different algorithms. 

\begin{figure}[h]
    \centering
    \begin{subfigure}[b]{0.45\textwidth}
    \centering
        \includegraphics[width=\linewidth]{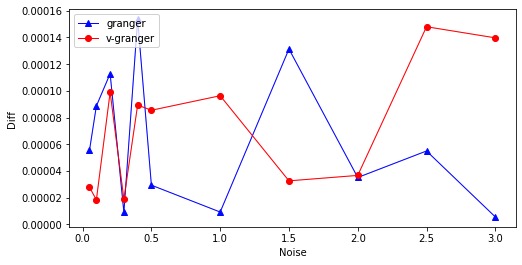}         
         \caption{Temperature data result under different noise level} 
         \label{fig:inference}
    \end{subfigure}
    \begin{subfigure}[b]{0.45\textwidth}  
    \centering
        \includegraphics[width=\linewidth]{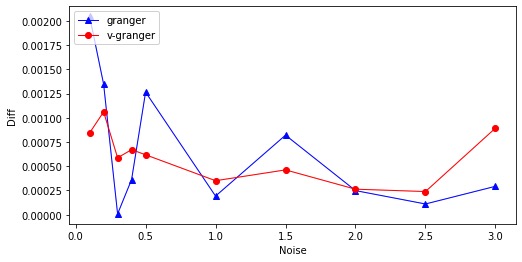} 
        \caption{Milk data result under different noise level} 
        \label{fig:generative}
    \end{subfigure}
       \caption{The result for the two Granger test method under different noise level of proxy. The Y axis is the $Diff(\vec{m})=|GC(\vec{x}\to\vec{y}|\vec{m})-GC(\vec{x}\to\vec{y}|\vec{z})|$. $\vec{m}$ is $\vec{p}$ for granger.  $\vec{m}$ is estimated $\vec{z}$ for \dname{}. The dimensionality of $\vec{z}$ is 1.}
    \label{fig:noise_result}
\end{figure}

\begin{figure}[h]
    \centering
    \begin{subfigure}[b]{0.45\textwidth}
    \centering
        \includegraphics[width=\linewidth]{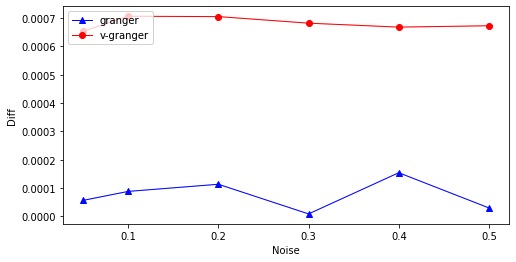}         
         \caption{Temperature data result under different noise level} 
         \label{fig:inference}
    \end{subfigure}
    \begin{subfigure}[b]{0.45\textwidth}  
    \centering
        \includegraphics[width=\linewidth]{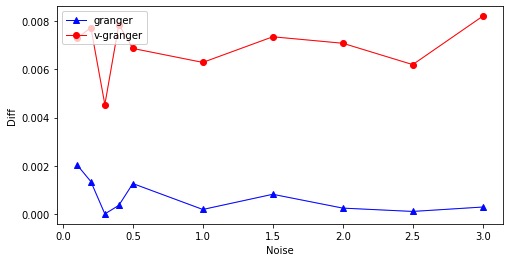} 
        \caption{Milk data result under different noise level} 
        \label{fig:generative}
    \end{subfigure}
    \caption{The dimensionality of $\vec{z}$ is 3.}
       
    \label{fig:noise_result_dimz_3}
\end{figure}

For each time series in these two datasets, there is only one sample. So we adopt the sliding window to get the training samples. For Butter dataset, the window length is 100. For the SML2010, the window length is 400. We use a 1-dimensional continuous estimated $\vec{z}$ in the \dname{}. The result is shown in figure \ref{fig:noise_result}. {\color{red} We can see from figure \ref{fig:noise_result}, \dname{} is better than the Granger test condition on the $\vec{p}$ in some noise levels, but the performance is worse with the noise level growth. We tried to adjust the parameter the dimensionality of estimated $\vec{z}$, the results didn't get better(shown in figure \ref{fig:noise_result_dimz_3}). } 

\begin{figure}[h]
    \centering
    \begin{subfigure}[b]{0.45\textwidth}
    \centering
        \includegraphics[width=\linewidth]{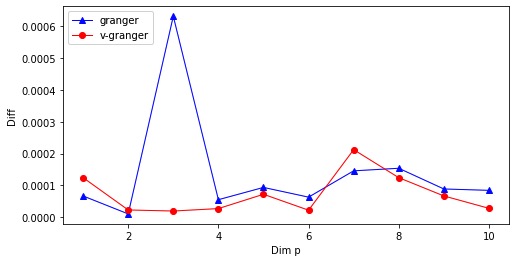}         
         \caption{Temperature data result under different noise level} 
         \label{fig:inference}
    \end{subfigure}
    \begin{subfigure}[b]{0.45\textwidth}  
    \centering
        \includegraphics[width=\linewidth]{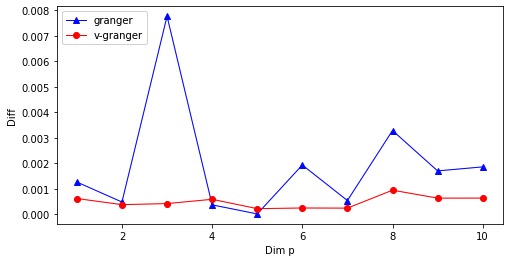} 
        \caption{Milk data result under different noise level} 
        \label{fig:generative}
    \end{subfigure}
      \caption{The result for the two Granger test method under different number of proxy. The Y axis is the $Diff(\vec{m})=|GC(\vec{x}\to\vec{y}|\vec{m})-GC(\vec{x}\to\vec{y}|\vec{z})|$. $\vec{m}$ is $\vec{p}$ for granger. The X axis is the dimensionality of the proxies. The noise level of each proxy is 0.5. $\vec{m}$ is $\hat{\vec{z}}$ for \dname{}.}    \label{fig:dimp}
\end{figure}

We also compare the performance of the traditional Granger test and the \dname{} under different number of the $\vec{p}$, the results are shown in figure \ref{fig:dimp}. \dname{} is better than the traditional Granger.


\section{Conclusion}
\label{section:conclusion}
Here we list the main problems of our paper.
\begin{itemize}

\item We lack a non-linear conditional Granger test method which is not sensitive to the dimensionality of the conditional variable.

\item The performance of our method is worse than the proxy variable in some cases. It is not clear whether our method is adapted to this problem. How can we improve the performance?
\item The real dataset with the truth proxy variable is need to evaluate our method.

\end{itemize}

\clearpage
\bibliography{reference}
\bibliographystyle{aaai}

\end{document}